\documentclass[ASNA,twocolumn]{USG} 
\usepackage{anyfontsize} %

\usepackage{colortbl}   
\usepackage{xcolor}     

\usepackage{steinmetz}
\graphicspath{{./images/}}
\specialissue{AE-CAI | CARE | OR 2.0 | PRiSM of Wiley's IET Healthcare Technology Letters}

\articletype{ORIGINAL ARTICLE}%
\subarticletype{Workshop Proceedings}

\received{16 July 2026}
\revised{x}
\accepted{x}
\journal{Wiley's IET Healthcare Technology Letters}
\volume{0}
\copyyear{202x}
\startpage{1}
\articledoi{10.1002/0000}

\begin{document}
\title{Looking Beyond the Scale: Do Surgical Skill Models Learn Transferable Representations Across Assessment Rubrics?}
\transtitle{Looking Beyond the Scale: Do Surgical Skill Models Learn Transferable Representations Across Assessment Rubrics?}

\author[1,2,3]{Hanna Hoffmann}[https://orcid.org/0009-0008-4917-7911]
\author[4,5]{Felix von Bechtolsheim}
\author[1,6,7]{Stefanie Speidel}
\author[1,2]{Rebecca Hisey}

\authormark{HOFFMANN \textsc{et al.}}

\address[1]{\orgdiv{Department of Translational Surgical Oncology}, \orgname{National Center for Tumor Diseases (NCT), NCT/UCC Dresden, a partnership between DKFZ, Faculty of Medicine and University Hospital Carl Gustav Carus, TUD Dresden University of Technology, and Helmholtz-Zentrum Dresden-Rossendorf (HZDR)}, \orgaddress{\country{Germany}}}

\address[2]{\orgdiv{Faculty of Medicine and University Hospital Carl Gustav Carus}, \orgname{Technische Universität Dresden (TUD)}, \orgaddress{\state{Dresden}, \country{Germany}}}

\address[3]{\orgdiv{BMFTR Research Hub 6G-Life}, \orgname{Technische Universität Dresden (TUD)}, \orgaddress{\state{Dresden}, \country{Germany}}}

\address[4]{\orgdiv{Department of Visceral, Thoracic and Vascular Surgery}, \orgname{Medical Faculty and University Hospital Carl Gustav Carus, TUD Dresden University of Technology}, \orgaddress{\state{Dresden}, \country{Germany}}}

\address[5]{\orgdiv{Surgical Skills Lab Dresden}, \orgname{Medical Faculty and University Hospital Carl Gustav Carus, TUD Dresden University of Technology}, \orgaddress{\state{Dresden}, \country{Germany}}}

\address[6]{\orgdiv{Deutsches Krebsforschungszentrum (DKFZ)}, \orgaddress{\state{Dresden}, \country{Germany}}}

\address[7]{\orgdiv{The Centre for Tactile Internet with Human-in-the-Loop (CeTI)}, \orgname{Technische Universität Dresden (TUD)}, \orgaddress{\state{Dresden}, \country{Germany}}}

\corres{Hanna Hoffmann. \email{hanna.hoffmann@nct-dresden.de}}





\transkeywords{surgical skill, surgical training, surgical assessment, laparoscopic training, video analysis, skill assessment, GOALS, LASANA, OSATS, JIGSAWS, machine learning, deep learning, self supervised learning, contrastive learning}

\abstract[ABSTRACT]{Vision-based surgical skill assessment has shown strong in-domain results, yet a fundamental question remains unasked: do these models learn transferable representations of surgical proficiency, or do they merely encode dataset-specific visual patterns?

This paper systematically analyzes what limits cross-domain skill transfer between the GOALS and OSATS assessment scales using the LASANA and JIGSAWS datasets. Each evaluated method serves a targeted diagnostic purpose: end-to-end training to test whether supervised skill learning transfers directly, Adaptive Sharpness-Aware Minimization (ASAM) to probe whether flatter loss landscapes improve generalization, and augmentation-based self-supervised and contrastive learning to assess whether domain-invariant pretraining decouples skill from visual context. Transfer is evaluated in both directions using a disjoint-participant held-out test set for JIGSAWS.

Results reveal an asymmetry: backbones pretrained on JIGSAWS achieve CCC values of 0.77 to 0.80 on LASANA, closely matching the end-to-end baseline, showing cross-rubric transfer is feasible when the target domain provides consistent supervision. Transfer to JIGSAWS fails across all methods, likely due to annotation inconsistencies. Control experiments with a Kinetics-pretrained backbone suggest task-specific heads carry the majority of the skill prediction burden, while the backbone need only provide adequate spatiotemporal features.

These findings offer a new perspective on vision-based skill assessment: the central question of whether skill representations transfer across scoring systems has not been previously investigated. Results indicate the visual component is dominant but not solely responsible for skill prediction; further work is needed to conclusively disentangle transferable skill features from those bound to a specific visual domain.}

 \transabstract[transABSTRACT]{Vision-based surgical skill assessment has shown strong in-domain results, yet a fundamental question remains unasked: do these models learn transferable representations of surgical proficiency, or do they merely encode dataset-specific visual patterns?

This paper systematically analyzes what limits cross-domain skill transfer between the GOALS and OSATS assessment scales using the LASANA and JIGSAWS datasets. Each evaluated method serves a targeted diagnostic purpose: end-to-end training to test whether supervised skill learning transfers directly, Adaptive Sharpness-Aware Minimization (ASAM) to probe whether flatter loss landscapes improve generalization, and augmentation-based self-supervised and contrastive learning to assess whether domain-invariant pretraining decouples skill from visual context. Transfer is evaluated in both directions using a disjoint-participant held-out test set for JIGSAWS.

Results reveal an asymmetry: backbones pretrained on JIGSAWS achieve CCC values of 0.77 to 0.80 on LASANA, closely matching the end-to-end baseline, showing cross-rubric transfer is feasible when the target domain provides consistent supervision. Transfer to JIGSAWS fails across all methods, likely due to annotation inconsistencies. Control experiments with a Kinetics-pretrained backbone suggest task-specific heads carry the majority of the skill prediction burden, while the backbone need only provide adequate spatiotemporal features.

These findings offer a new perspective on vision-based skill assessment: the central question of whether skill representations transfer across scoring systems has not been previously investigated. Results indicate the visual component is dominant but not solely responsible for skill prediction; further work is needed to conclusively disentangle transferable skill features from those bound to a specific visual domain.}




\copyright{This is an open access article under the terms of the \href{Creative Commons Attribution-NonCommercial}{Creative Commons Attribution-NonCommercial} License, which permits use, distribution and reproduction in any medium, provided the
original~work~is~properly cited and is not used for commercial purposes.
\\[5pt]
  ©  2024 The Author(s) \textit{AIChE Journal} published by Wiley Periodicals LLC on behalf of American Institute of Chemical Engineers.}


\maketitle


\section{Introduction}\label{sec:intro}

The landscape of surgical education has shifted considerably from its traditional apprenticeship roots, where trainees acquired competence primarily through observation and supervised repetition in the operating room~\cite{Seymour2002}. Although this model has been in use for centuries, it is limited by inconsistent exposure and the subjectivity of the evaluators, and therefore has recently been the subject of debate regarding its ethical implications and patient safety. The integration of simulation-based curricula has helped address some of these shortcomings~\cite{Elendu2024}, offering repeatable, low-risk environments for deliberate practice. Concurrently, structured assessment instruments such as the Objective Structured Assessment of Technical Skills (OSATS)~\cite{Martin1997} and the Global Operative Assessment of Laparoscopic Skills (GOALS)~\cite{Vassiliou2005}, were introduced to reduce evaluator bias and provide criterion-referenced feedback. Yet even with standardized rubrics, manual assessment remains resource-intensive: each evaluation requires dedicated expert time, limiting the frequency and scalability of feedback that trainees receive. Given the established link between technical proficiency and patient outcomes~\cite{Birkmeyer2013}, there is strong motivation to develop automated methods that can deliver objective, timely skill assessment without placing additional burden on faculty.

Two commonly used rating scales are OSATS~\cite{Ahmed2011,Martin1997}, developed for open surgery but since adapted to other contexts, and GOALS~\cite{Vassiliou2005}, introduced for laparoscopic surgery. While both capture technical proficiency, they differ in scale, granularity, and assessed competencies. Prior work has largely treated these as independent systems, with models trained and evaluated within a single framework~\cite{Lam2022,Levin2019}. Whether skill representations learned under one rubric transfer to another remains unexplored, yet serves as a natural probe into whether models encode transferable notions of proficiency or merely rubric- and domain-specific visual patterns.

Current automated approaches~\cite{Lam2022,Levin2019,Pedrett2023} span a range of modalities. Tool motion data, extracted either from robot kinematics~\cite{Benmansour2023,Ogul2022} or video-based tracking~\cite{Fathabadi2021,Lavanchy2021,Lazar2023}, has been widely used to compute trajectory-based features indicative of skill. More recently, vision-based deep learning methods have gained traction, operating directly on raw surgical video to extract visual and temporal features for skill prediction~\cite{Anastasiou2023,Hoffmann2024,Kiyasseh2023-tf}. These approaches are particularly appealing as they require no additional hardware or instrumentation, making them applicable across robotic, laparoscopic, and potentially open surgical settings~\cite{Hamza2025}.

Despite their promise, vision-based models are typically developed and evaluated within a single assessment framework and surgical context. A model trained on robotic tasks may learn features that are specific to the visual characteristics of that domain--the appearance of robotic instruments, the simulation environment, or the particular task geometry--rather than generalizable representations of surgical proficiency. This raises a fundamental unaddressed question: do vision-based skill models learn transferable features that capture underlying technical competence, or do they primarily encode domain-specific visual patterns tied to a particular scoring rubric and surgical setting? Understanding this distinction is critical for developing scalable assessment tools that generalize across procedures, environments, and evaluation frameworks. This is a prerequisite for eventual translation to clinical settings, where visual variability in anatomy, instrumentation, and procedures is the norm.

\begin{figure*}
\centerline{\includegraphics[width=\textwidth]{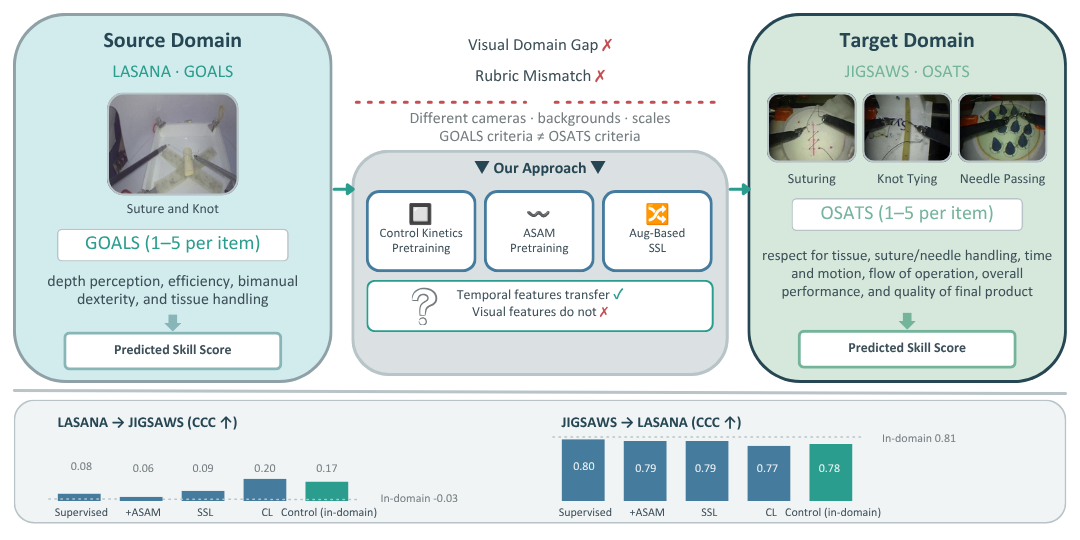}}
\caption{Overview of the cross-domain surgical skill transfer problem and our approach. JIGSAWS (annotated with OSATS) and LASANA (annotated with GOALS) differ substantially in visual appearance and scoring rubric. This work investigates whether vision-based skill representations transfer across these domains using image normalization, ASAM pretraining, and augmentation-based self-supervised learning.\label{fig1}}
\end{figure*}

In this work, summarized in Figure~\ref{fig1}, we pose a question that has, to our best knowledge, not been addressed in the surgical skill assessment literature: what do vision-based models actually learn about skill? This is approached by using an unexplored cross-rubric transfer between GOALS and OSATS as a diagnostic tool. If models capture genuine, transferable notions of proficiency, representations learned under one rubric should exhibit meaningful predictive power under another that assesses related constructs. Using the JIGSAWS~\cite{jigsaws} and LASANA~\cite{lasana} datasets, which differ substantially in visual appearance, surgical context, and annotation scheme, our analysis evaluates transfer in both directions and probes the relative contribution of visual versus more abstract features to skill prediction under domain shift. We further adopt a disjoint-participant held-out test set for JIGSAWS evaluation, providing a stricter generalization setting than the cross-validation protocols commonly used in the literature~\cite{Ahmidi2017,Anastasiou2023, Hendricks2024}. To our knowledge, this represents the first multitask evaluation on JIGSAWS under this more stringent generalization setting. To encourage more transferable representations, we explore Adaptive Sharpness-Aware Minimization (ASAM) during pretraining and augmentation-based self-supervised and contrastive learning for surgical skill assessment.

\section{Related Work}\label{sec:relwork}

\subsection{Vision-Based Surgical Skill Assessment}

Early automated skill assessment methods relied heavily on kinematic data from robotic systems or external sensors, with JIGSAWS~\cite{Ahmidi2017} serving as the predominant benchmark for kinematic-based approaches. More recently, vision-based methods have emerged that operate directly on surgical video, removing the dependency on instrumented environments. Anastasiou et al.~\cite{Anastasiou2023} and Funke et al.~\cite{Funke2019} demonstrated that deep learning models can extract meaningful skill-relevant features from video alone, achieving competitive performance with kinematic approaches on robotic tasks. Furthermore, AIxSuture~\cite{Hoffmann2024} extended vision-based skill prediction to open surgery, demonstrating the feasibility of video-only assessment beyond the robotic domain.

The temporal dimension of surgical video has proven particularly informative for skill discrimination. Models incorporating temporal modeling consistently outperform frame-level approaches~\cite{Hoffmann2026}, suggesting that skill manifests not only in individual frames but in the sequential structure of surgical actions.

\subsection{Domain Adaptation and Transfer Learning in Surgical AI}

In general, domain shift is a well-documented challenge in surgical computer vision, arising from differences in anatomy, instrumentation, camera viewpoints, and institutional practices. While domain adaptation techniques have been applied extensively to tasks such as surgical phase recognition and instrument segmentation, their application to skill assessment remains limited.

Luongo et al.~\cite{Wang2023} explored uncertainty-aware self-supervised learning (SSL) for cross-domain technical skill assessment in robot-assisted surgery, though their approach relied on kinematic data and was evaluated only in simulated environments. Goldbraikh et al.~\cite{Gong2025} evaluated the generalizability of video-based skill assessment in ophthalmology, finding limited transfer to new clinical contexts. More recently, Anastasiou et al.~\cite{Anastasiou2025} explored pre-training across domains for few-shot surgical skill assessment, demonstrating the challenges of learning transferable representations under limited data.

Most notably, Anastasiou et al.~\cite{Anastasiou2026} introduced CoRe-DA, the first benchmark for unsupervised domain adaptation (UDA) in surgical skill assessment regression. Their work spans four datasets across dry-lab and clinical settings, defining two UDA settings: AIxSuture (OSATS) to JIGSAWS (OSATS), and RAH-skill (M-GEARS) to RARP-skill (M-GEARS). CoRe-DA employs contrastive regression to learn domain-invariant representations through relative score supervision and target domain self-training, achieving Spearman Correlation Coefficients of 0.46 and 0.41 on target datasets without any labeled target data~\cite{Anastasiou2026}. Critically, their results reveal that existing state-of-the-art skill assessment models generalize poorly under domain shift, with source-only baselines achieving near-zero correlation on target domains. This confirms that current vision-based models learn representations that are largely domain-specific.

Our work differs from CoRe-DA in two key respects. First, we investigate transfer across different assessment \textit{rubrics} (GOALS $\leftrightarrow$ OSATS) rather than within the same label space, introducing an additional layer of semantic misalignment beyond visual domain shift. Second, we focus on understanding \textit{what} vision-based models learn (visual vs. temporal features) and how pretraining strategies (SSL, ASAM) can encourage more transferable representations, rather than proposing a new framework.

\subsection{Self-Supervised Learning for Surgical Video}

SSL has gained traction as a strategy for learning robust representations from unlabeled surgical video, which is abundantly available but expensive to annotate with skill labels. Augmentation-based contrastive methods encourage models to learn features invariant to visual perturbations, potentially disentangling skill-relevant temporal patterns from domain-specific visual characteristics. The CoRe-DA framework itself leverages contrastive principles combined with background mixing to partially align low-level visual statistics across domains~\cite{Anastasiou2026}, and their ablation studies confirm that this visual normalization contributes to adaptation performance.

However, the specific utility of augmentation-based SSL pretraining for bridging the gap between different assessment domains, where both visual appearance and the scoring rubric itself differ, has not been systematically studied. To our knowledge, no prior work has examined whether SSL can help vision models learn representations that transfer across GOALS and OSATS evaluation frameworks. This gap motivates our exploration of SSL as a pretraining strategy for cross-rubric transfer.

\section{Methodology}\label{sec:method}
\subsection{Datasets}
Two datasets are used in this work: JIGSAWS and LASANA.

\paragraph*{JIGSAWS}~ The JHU-ISI Gesture and Skill Assessment Working Set (JIGSAWS)~\cite{jigsaws} is a publicly available surgical activity dataset collected on the da Vinci Surgical System. It contains synchronized stereo video and robot kinematic data from eight participants with varying levels of surgical experience, categorized as Novice, Intermediate, or Expert based on self-reported experience.

Participants performed three benchtop surgical tasks (suturing, needle passing, and knot tying) with up to five repetitions each, resulting in 103 videos in total (39 suturing, 36 knot tying, and 28 needle passing). The mean video duration varies by task, at approximately 1 minute 53 seconds for suturing, 57 seconds for knot tying, and 1 minute 48 seconds for needle passing . Each video is annotated video-level skill assessment based on the OSATS rating scale, excluding items that were not applicable to the surgical tasks. The OSATS-based rating comprises six categories (respect for tissue, suture/needle handling, time and motion, flow of operation, overall performance, and quality of final product), each scored on a Likert scale from 1 to 5. The modified GRS is also provided and is computed as the sum of all six category scores, yielding a range of 6 to 30.

The JIGSAWS dataset defines two standard cross-validation schemes: Leave-One-Supertrial-Out (LOSO) and Leave-One-User-Out (LOUO). In LOSO, all trials sharing the same repetition index across subjects are grouped into a single "supertrial," yielding a 5-fold evaluation that tests generalization across practice attempts while still exposing the model to all subjects during training. In LOUO, each of the eight folds holds out all recordings from one subject entirely, providing a more stringent evaluation of generalization to previously unseen users. Since an untouched test set was required for JIGSAWS, the standard predefined cross-validation schemata could not be adopted. Instead, the dataset is divided by participant and stratified by quantile-binned mean GRS scores. Three bins are used, and the resulting splits follow a 70/10/20 ratio (train/validation/test) incorporating data from all tasks, ensuring that there is no participant overlap between the training, validation, and test sets.

\paragraph*{LASANA}~ The Laparoscopic Skill Analysis and Assessment (LASANA)~\cite{lasana} dataset is a large-scale benchmark for video-based surgical skill analysis, comprising 1270 stereo video recordings of four basic laparoscopic training tasks (peg transfer, circle cutting, balloon resection, and suture and knot) performed in a laparoscopic training box. The dataset includes recordings from 70 participants (58 medical students and 12 clinicians) with varying levels of laparoscopic experience. Medical students were recorded at multiple stages throughout a laparoscopic training course, contributing up to six recording sessions per task, while clinicians each performed a single session . For the suture and knot task, which is the only task used in this work, 314 videos are available with a mean duration of 4 minutes 30 seconds. Each recording is annotated with modified GOALS skill rating developed by Kowalewski et al.~\cite{Kowalewski2016}, assessing four dimensions (depth perception, efficiency, bimanual dexterity, and tissue handling) on a five-point Likert scale. The total GRS is computed as the sum across all four dimensions. To enhance reliability, each video was independently rated by four assessors, and the final scores were obtained by averaging the maximum-normalized ratings.

Out of the four LASANA tasks, only the suture and knot task is used because it is the visually closest to the three JIGSAWS tasks, all of which comprise elements of suturing and thread manipulation. This choice reflects the goal of focusing on skill domain transfer rather than visual domain transfer. Although intrinsic visual differences between LASANA and JIGSAWS remain present due to differences in the surgical platform (laparoscopic training box versus da Vinci Surgical System), selecting the most visually similar task minimizes confounding visual domain shift as much as possible.
For LASANA, the predefined dataset splits are used (232 training, 32 validation, and 50 test recordings for the suture and knot task).

\subsection{Model}
The proposed approach builds upon the state-of-the-art single-task surgical skill assessment model introduced by Funke et al.~\cite{lasana}, which combines a spatiotemporal 3D CNN backbone (X3D)~\cite{feichtenhofer2020} with a simple temporal aggregation scheme. The backbone is pretrained on the Kinetics-400 dataset~\cite{carreira2017quo}.

To enable model assessment across rating scales, the model is extended from single-task to multitask learning. The architecture is adapted with multiple prediction heads to simultaneously predict the different scoring categories within the GOALS or OSATS rubric. The number of heads is adjusted depending on the dataset used for training, and the combined Global Rating Scale (GRS) score is computed as the sum of the individual category scores. Additionally, an LSTM layer is introduced before the average pooling clip aggregation scheme in the model head, enabling the capture of temporal dependencies across clips. The full architecture follows this pipeline: X3D backbone, LSTM, clip aggregation via average pooling, and a multi-head linear layer for each scoring category.

During training, videos are sampled clip-wise by first subdividing each video into equal subsections and then randomly sampling one clip from each subsection. Each clip is 16 frames in length, with every 5th frame selected. Frames are center-cropped to 224×224 pixels and normalized to the mean and standard deviation of the ImageNet dataset~\cite{deng2009imagenet}. The model is trained on the source dataset and evaluated on the target dataset.

\subsubsection*{Workflow}
The same model architecture is used for both the supervised and unsupervised learning approaches. The primary difference lies in how the models are trained. Once training is complete, the prediction heads are adapted to the target dataset, if applicable, and finetuned.
\paragraph*{Supervised}~
In the supervised setting, the model is trained end-to-end using a Huber loss function. The loss is calculated for each head separately, and the total loss is computed as the sum of the individual losses.
\paragraph*{Unsupervised}~
In the unsupervised setting, the model is trained in two stages. First, the backbone is pretrained using either self-supervised learning (SSL) or contrastive learning (CL). Then, the model heads are finetuned on the target dataset using supervised learning, with the backbone weights frozen during this stage. The model is subsequently evaluated on the target dataset.
For self-supervised backbone training, the VICReg loss is used in combination with augmentation-based SSL, where augmented views of the same sample serve as training signal. These  augmentations include random cropping, horizontal flipping, affine, color jitter, gray scale, gaussian blur, and erasing. For contrastive backbone training, a triplet margin loss is employed. This approach also leverages augmentation-based SSL; however, in this case, the positive sample is generated through augmentation of the anchor, while the negative sample, which stems from a different video, is not augmented.

\section{Experiments and Results}\label{sec:exp}
The experiments investigate various approaches to enable transfer learning between the two surgical skill assessment scales, GOALS and OSATS. All studies are benchmarked against the multitask baseline model, and the same data splits are used consistently throughout to ensure fair comparison. Model performance is evaluated using the Concordance Correlation Coefficient (CCC) and Mean Absolute Error (MAE). The CCC is selected as the primary metric because it captures both correlation and agreement between predicted and true skill levels, making it well-suited for reflecting the ordinal nature of surgical proficiency ratings. MAE is reported alongside CCC to provide an intuitive measure of absolute prediction error magnitude and to facilitate comparison with prior work in the automated assessment community

\begin{center}
\begin{table*}[!ht]%
\caption{This table shows the test set results of the models pretrained on the LASANA datset. The middle and right columns show the results of the finetuned heads using the respective dataset in the column header. All values are given as the mean of three trials with the standard deviation in parentheses. \label{tab:lasana}}
\begin{tabular*}{\textwidth}{@{\extracolsep\fill}lllll@{}}
\toprule
&\multicolumn{2}{@{}l}{\textbf{LASANA}} & \multicolumn{2}{@{}l}{\textbf{JIGSAWS}} \\\cmidrule{2-3}\cmidrule{4-5}
\textbf{Model} & \textbf{CCC}  & \textbf{MAE}  & {\textbf{CCC}}  & \textbf{MAE}   \\
\midrule
Kinetics & 0.78 (0.03) & 0.07 (0.007) & - & - \\
Supervised & \textbf{0.81 (0.01)$^{\tnote{*}}$}  & \textbf{0.07 (0.004)$^{\tnote{*}}$}  & 0.08 (0.08) & 0.16 (0.011)   \\
ASAM & 0.74 (0.04)$^{\tnote{*}}$  & 0.07 (0.004)$^{\tnote{*}}$  & 0.06 (0.04)  & 0.18 (0.012)   \\
SSL & 0.78 (0.06)  & 0.07 (0.012)  & 0.09 (0.09)  & 0.19 (0.006)   \\
CL & 0.80 (0.01)  & 0.07 (0.001)  & 0.20 (0.06)  & 0.16 (0.004)   \\
\bottomrule
\end{tabular*}
\begin{tablenotes}
\item[$^{*}$] End to end trained model.
\end{tablenotes}
\end{table*}
\end{center}

\begin{center}
\begin{table*}[!ht]%
\caption{This table shows the test set results of the models pretrained on the JIGSAWS datset. The middle and right columns show the results of the finetuned heads using the respective dataset in the column header. All values are given as the mean of three trials with the standard deviation in parentheses. \label{tab:jigsaws}}
\begin{tabular*}{\textwidth}{@{\extracolsep\fill}lllll@{}}
\toprule
&\multicolumn{2}{@{}l}{\textbf{JIGSAWS}} & \multicolumn{2}{@{}l}{\textbf{LASANA}} \\\cmidrule{2-3}\cmidrule{4-5}
\textbf{Model} & \textbf{CCC}  & \textbf{MAE}  & {\textbf{CCC}}  & \textbf{MAE}   \\
\midrule
Kinetics & 0.17 (0.01) & 0.19 (0.06) & - & - \\
Supervised & \textbf{-0.03 (0.08)$^{\tnote{*}}$}  & \textbf{0.23 (0.012)$^{\tnote{*}}$}  & 0.80 (0.01) & 0.07 (0.002)   \\
ASAM & 0.22 (0.01)$^{\tnote{*}}$  & 0.18 (0.009)$^{\tnote{*}}$  & 0.79 (0.05)  & 0.07 (0.004)   \\
SSL & 0.22 (0.05)  & 0.18 (0.012)  & 0.79 (0.02)  & 0.07 (0.004)   \\
CL & 0.05 (0.02)  & 0.18 (0.003)  & 0.77 (0.04)  & 0.07 (0.006)   \\
\bottomrule
\end{tabular*}
\begin{tablenotes}
\item[$^{*}$] End to end trained model.
\end{tablenotes}
\end{table*}
\end{center}

\paragraph*{Baseline}~
Two baselines are established for comparison. First, an in-domain baseline is obtained by training the multitask model end-to-end in a fully supervised manner on a given dataset and evaluating on the same dataset. These results are shown in bold in Tables~\ref{tab:jigsaws} and~\ref{tab:lasana} (Supervised, left columns) and represent the upper-bound reference for same-domain performance. Second, a cross-domain baseline is established by taking the same trained model, replacing the prediction heads to match the target dataset categories, and finetuning only the heads on the target dataset. These results are reported in the right columns of Tables~\ref{tab:jigsaws} and~\ref{tab:lasana} (Supervised row) and represent the naive transfer performance without any domain adaptation applied to the backbone. The cross-domain baseline serves as the primary reference point against which subsequent transfer techniques are compared.

For further comparison, the model is also evaluated with just a Kinetics-400 pretrained backbone and fine-tuned end-to-end trained heads on the target dataset. This allows for an assessment of how much the model benefits from supervised and unsupervised pretraining on the source dataset versus relying solely on generic video features in the backbone learned from Kinetics-400.

The baseline achieves a CCC of 0.81 and -0.03 and MAE of 0.07 and 0.23 on LASANA AND JIGSAWS, respectively, for the in-domain target training and evaluation. The LASANA results align with results from the original study. However, results on JIGSAWS are shockingly low.

To address whether the underlying issue was with the model or the dataset, the model was extensively tested with different hyperparameters including loss, early stopping schemes, normalization augmentations -- even leaving out aumentations altogether, seeds, splits, and learning rates. Model alterations were also tested by excluding the additional duration features and excluding the LSTM layers in the head. All trials resulted in similar results on the JIGSAWS dataset, leaving us to assume that the issue seems to be an underlying aspect of the JIGSAWS dataset itself. This is analysed further in the Discussion section (\ref{sec:disc}). Therefore, the results and their interpretation must be regarded critically within this context.

\paragraph*{Supervised Training Study}~
To investigate whether minimal interventions can improve cross-domain transfer without major architectural changes,  Adaptive Sharpness-Aware Minimization (ASAM) is employed during training. ASAM has been shown to improve generalization across heterogeneous data distributions in federated learning settings \cite{Caldarola2022}. The hypothesis is that training toward flat minima encourages the model to learn more generalizable visual representations, thereby improving transfer to a visually distinct domain and, by extension, to a different assessment scale.
 
When predicting in domain on LASANA as well as for transfering between either datatset, the ASAM model's performance is slightly lower than the Supervised model. For the in-domain performance, this is likely attributable to the flatter minima optimization provoked by the ASAM technique, causing the model to not quite reach optimum performance with hopes of better generalizability. Overall transfer remains limited; the addition of ASAM does not seem to give much of a benefit.

The supervised ablation results show that ASAM does not meaningfully improve transfer performance in either direction, suggesting that flat minima optimization alone is insufficient to address the domain gap between the two scales. Though conclusions from JIGSAWS evaluation are limited by annotation inconsistencies in the dataset. Given that lightweight training-time interventions such as ASAM do not address the observed transfer asymmetry, and that the challenging transfer direction (LASANA to JIGSAWS) may benefit from stronger representation learning, this motivates the exploration of pretraining strategies that explicitly encourage the learning of domain-agnostic visual representations.

\paragraph*{Self-Supervised and Contrastive Pretraining}~
The hypothesis for this set of experiments is that by learning visual representations that are invariant to appearance differences through pretraining, the model can better disentangle domain-specific visual characteristics from skill-relevant features. Self-supervised pretraining has been shown to produce more transferable and domain-agnostic representations, as the learned features are not tied to task-specific labels from a single domain \cite{Ericsson2021,He2020}. When subsequently finetuned for skill assessment on the target dataset, such a model may more effectively focus on the skill domain rather than being confounded by visual domain shift.

Two distinct pretraining methods are compared. Pure augmentation-based SSL using VICReg loss has not yet been extensively explored in the surgical skill assessment domain, offering a novel perspective. Augmentation-based contrastive learning using triplet margin loss builds on existing research in skill assessment and visual domain transfer but introduces a twist through its augmentation strategy.

A seen in Table~\ref{tab:lasana}, the in-domain CL approach performs on par with the end-to-end trained Supervised model. When transfering from LASANA to JIGSAWS, both SSL and CL methods achieve better results than the fully supervised methods.

In Table~\ref{tab:jigsaws}, in-domain evaluation seems to perform better than the baseline with SSL achieving significantly better results than both the Supervised and the CL approaches. However as stated previously, these interpretations must be considered with care. When transfering to LASANA, both SSL and CL perform below the Supervised cross-domain model as well as the in-domain models for LASANA. Although, the SSL JIGSAWS-LASANA cross-domain model has a slightly better performance than the in-domain LASANA SSL model.

\section{Discussion}\label{sec:disc}
The central finding of these results is a pronounced asymmetry in cross-dataset skill transfer most likely attributable to the discrepancies within the JIGSAWS dataset. Backbones pretrained on JIGSAWS and transferred to LASANA achieve CCC values between 0.77 and 0.80 (Table~\ref{tab:jigsaws}, right columns), closely matching the LASANA end-to-end Supervised baseline of 0.81 (Table~\ref{tab:lasana}). This consistency holds across all four methods, with standard deviations of 0.01 to 0.05, indicating that the transfer is both effective and stable regardless of pretraining strategy. Notably, even the Supervised model, trained end-to-end on only $\sim$70 JIGSAWS videos, produces a backbone that supports near-baseline skill prediction when transfering to LASANA. These results suggests that spatiotemporal features relevant to surgical skill can be extracted from a small robotic surgery dataset and applied successfully in a visually distinct laparoscopic domain, provided the target dataset offers sufficient well-annotated data for head training. However, this interpretation must be considered with care.

The reverse direction produces a starkly different outcome. Transfer from LASANA to JIGSAWS (Table~\ref{tab:lasana}, right columns) yields CCC values of only 0.06 to 0.20. At first glance, this asymmetry might suggest that features learned on laparoscopic box trainer data simply do not generalize to robotic surgery. However, inspection of the JIGSAWS end-to-end baselines (Table~\ref{tab:jigsaws}, left columns) reveals that this interpretation is insufficient. The Supervised model trained and evaluated entirely on JIGSAWS achieves a CCC of -0.03 ($\pm$ 0.08), indicating performance below chance level. Even the best-performing baseline methods (ASAM and SSL) reach only a CCC of 0.22. The inability to predict OSATS/GRS scores is therefore not specific to the transfer condition but reflects a fundamental difficulty in learning meaningful skill representations from JIGSAWS under a held-out test set protocol as used in this work.

A further observation reinforces this point. MAE values on JIGSAWS remain relatively compressed across all conditions (0.16 to 0.19 for transfer, 0.18 to 0.23 for baselines), while CCC values fluctuate. Since CCC captures both correlation and agreement in relative ordering while MAE reflects only absolute error magnitude, this pattern suggests that models can approximate the general range of scores but fail to reproduce their rank structure. This is precisely what would be expected if the ground truth annotations themselves lack a consistent ordering with respect to the visual features that encode actual skill differences.

The consistently poor results on JIGSAWS across all conditions, both transfer and baseline, raised the question of whether the cause was model-related or dataset-related. Extensive testing was conducted, including hyperparameter searches, architectural modifications, and training protocol adjustments. None of these interventions produced meaningful improvement, effectively ruling out model-side explanations. This prompted a closer examination of the JIGSAWS annotation structure itself.

\begin{figure}
\centerline{\includegraphics[width=78mm]{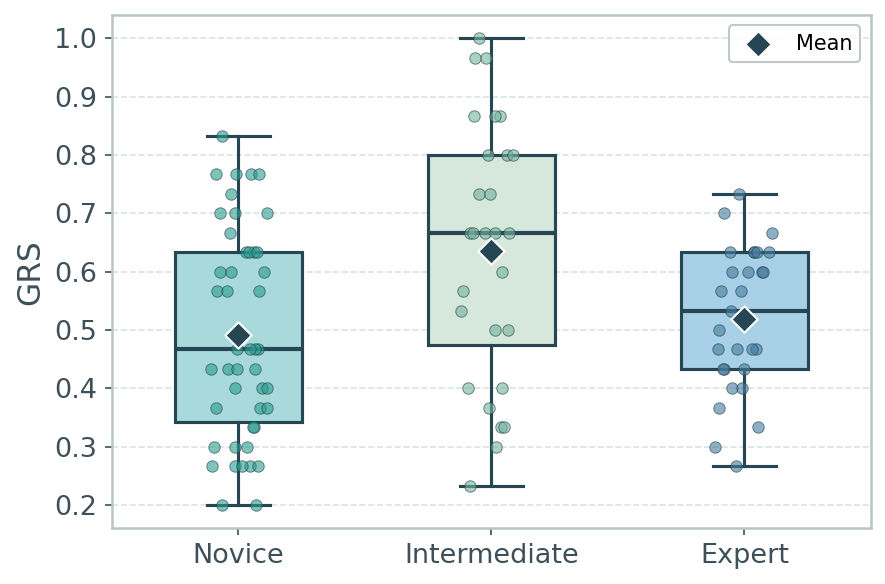}}
\caption{Distributions of the maximum normalized GRS compared to the self-proclaimed skill level based on the number of hours of experience. Boxplots are divided by quartiles; the mean is separately plotted.\label{fig:grsDist}}
\end{figure}

A review of individual JIGSAWS annotations revealed troubling inconsistencies: certain videos from participants classified as experts receive unexpectedly low GRS and OSATS scores, while some novice-level participants achieve scores comparable to or exceeding those of more experienced operators. Overall, the distributions of GRS do not reflect a clear modularity in skill, as observed in Figure~\ref{fig:grsDist}. These patterns directly explain the poor CCC values observed across all JIGSAWS conditions in this study: if the ground truth annotations lack consistent rank ordering with respect to actual skill, models cannot learn to reproduce such ordering regardless of backbone quality or pretraining strategy. The relatively stable MAE values (0.16 to 0.19 for transfer, 0.18 to 0.23 for baselines) compared to near-zero CCC values confirm this interpretation, as models approximate the general score range but fail to capture meaningful relative differences.

These observations are not isolated to the present work. Hendricks et al.~\cite{Hendricks2024} performed the first statistical analysis of the JIGSAWS dataset and found that operator performance is unrelated to training hours with high statistical significance (p < 0.005), with GRS distributions across NIE classes (Novice, Intermediate, Expert) showing substantial overlap and inconsistent ordering across tasks. Their analysis supports the hypothesis that prior works achieving high NIE classification accuracy under the typically reported LOSO protocols may have been recovering participant-specific patterns rather than generalizable skill representations. Critically, however, Hendricks et al. also found that robot-side kinematic indicators, particularly acceleration profiles, do correlate with GRS across tasks. This suggests that meaningful skill variation exists among the participants, but the OSATS/GRS labels do not reliably capture it in a way that vision-based supervised learning can exploit under rigorous evaluation conditions.

JIGSAWS has served as a foundational benchmark for nearly a decade, enabling significant progress in automated surgical skill assessment through its public availability, standardized tasks, and multimodal data. The results presented here do not diminish its historical contributions. However, the standard LOSO cross-validation protocol may have masked these annotation issues by allowing models to exploit participant-specific patterns across repeated trials. The held-out, participant-disjoint protocol used in the present study removes this potential source of data leakage and exposes the underlying label limitations. This also means that absolute performance numbers reported here are not directly comparable to published JIGSAWS benchmarks; the focus remains on relative performance across transfer conditions. A thorough re-evaluation of the JIGSAWS annotations, potentially involving multiple independent raters and correlation with objective kinematic metrics, would be a valuable contribution but falls beyond the scope of this work.

Beyond the dataset discrepancies, further discussion points exist. The asymmetry in dataset size (103 vs. 314 videos; 8 vs. 70 participants) interacts with the annotation quality issue in a way that could have influence on the directionality of transfer success. When transferring to LASANA, even a backbone trained on JIGSAWS's small and potentially noisy data produces useful features, because the LASANA target provides sufficient well-structured training data from a diverse participant pool for the heads to learn stable GOALS predictions. The quality and scale of the target dataset effectively compensate for source limitations, evidenced by the tight clustering of all transfer methods around CCC 0.77 to 0.80 on LASANA (Table~\ref{tab:jigsaws}). Conversely, when transferring to JIGSAWS, even a backbone pretrained on the larger and better-structured LASANA corpus cannot overcome annotation inconsistencies in the target. The heads are fine-tuned on labels that may not reliably encode skill differences, and the small number of training samples available after the participant-disjoint split (approximately 70 videos from 5 participants) leaves minimal room for the heads to learn robust mappings even if the labels were perfectly consistent. The combination of small sample size and questionable label quality creates conditions under which it is difficult for a pretraining strategy to succeed.

Two additional factors complicate cross-dataset transfer, though they are difficult to fully disentangle from the annotation and sample size issues. The visual domains differ substantially: JIGSAWS captures the daVinci system's endoscopic view with robotic instruments, while LASANA presents standard laparoscopic instruments in a box trainer. Camera perspectives, instrument appearance, and background all differ. Additionally, OSATS and GOALS assess overlapping but distinct skill constructs: OSATS emphasizes categories such as Respect for Tissue and Overall Performance, while GOALS targets Depth Perception and Bimanual Dexterity. The strong JIGSAWS-to-LASANA transfer (CCC 0.77 to 0.80) demonstrates that neither the visual domain gap nor the skill concept mismatch poses an insurmountable barrier when the target dataset is adequate. The backbone appears to capture spatiotemporal patterns that seem to carry relevance across visual domains and assessment frameworks. Whether the reverse failure reflects these gaps or primarily the JIGSAWS annotation issues cannot be definitively resolved without access to a re-annotated or substantially expanded version of JIGSAWS.

SSL and contrastive learning (CL) were hypothesized to produce more domain-agnostic features by avoiding skill-label supervision during backbone training. The results offer partial support for this hypothesis. In the LASANA-to-JIGSAWS direction, CL achieves the highest transfer CCC (0.20 $\pm$ 0.06), more than doubling the performance of supervised methods (Supervised: 0.08, ASAM: 0.06). This suggests that label-free pretraining may reduce overfitting to source-domain annotation idiosyncrasies and produce representations that are somewhat more amenable to cross-domain head training. However, even this best-case transfer result remains far below useful predictive performance, constrained by the same JIGSAWS-side limitations addressed previously. Additionally given the uncertainties in the JGISAWS annotations, it cannot be definitively concluded whether these results are due to the architectural differences. In the JIGSAWS-to-LASANA direction, all methods converge to CCC values between 0.77 and 0.80, with no meaningful separation between supervised and self-supervised pretraining. This convergence indicates that when the target dataset is well-structured and sufficiently large, pretraining strategy matters less than the availability of quality target labels. The backbone need only provide a reasonable spatiotemporal feature space; the heads handle the skill-specific mapping.

A conceptual question remains regarding what these frozen backbones encode. If the backbone is not trained on skill labels, skill-discriminative capacity must either emerge incidentally from learning general spatiotemporal features that happen to correlate with skill (e.g., motion smoothness, economy of movement, tissue interaction) or be learned entirely by the task-specific heads. To disentangle these contributions, a control experiment was conducted using a backbone pretrained on Kinetics, a large-scale action recognition dataset with no surgical or skill-related content, with heads trained directly on each target dataset. This control achieves a CCC of 0.78 on LASANA (Table~\ref{tab:lasana}) and 0.17 (Table~\ref{tab:jigsaws}) on JIGSAWS. The LASANA result is particularly informative: a backbone with no exposure to surgical data produces skill predictions nearly indistinguishable from those achieved by JIGSAWS pretrained backbones (CCC 0.77 to 0.80, Table~\ref{tab:jigsaws}) and closely approaching the end-to-end Supervised baseline (CCC 0.81, Table~\ref{tab:lasana}). This indicates that the majority of skill-predictive capacity resides in the task-specific heads rather than the backbone, and that general spatiotemporal visual features learned from any sufficiently large video corpus provide a representational basis adequate for downstream skill assessment when the target dataset is well-structured and sufficiently annotated.

However, a subtle but consistent signal suggests that skill-supervised pretraining contributes something beyond generic visual encoding. The Supervised model pretrained end-to-end on JIGSAWS with skill labels achieves a CCC of 0.80 when transferred to LASANA, marginally outperforming the Kinetics control (CCC 0.78). While this difference is small, it suggests that explicit optimization on skill annotations shapes the backbone's feature space in ways that are at least partially transferable across datasets and assessment scales. End-to-end skill training may encourage the backbone to emphasize features beyond visual aspects, providing a modest but real advantage over representations trained purely on action classification. At the same time, the narrow margin between the Kinetics control and surgical-domain pretraining indicates that most of what a 3D CNN learns for generic action recognition already captures the motion dynamics relevant to skill discrimination, tempering the practical necessity of domain-specific pretraining.

These observations carry practical implications for the field. If task-specific heads carry the primary skill-learning burden and general spatiotemporal backbones provide a sufficient representational foundation, then expensive surgical-domain pretraining may not be strictly necessary for building skill assessment systems, provided sufficient well-annotated target data is available. This could lower the barrier to deploying automated skill assessment in new surgical domains where large-scale pretraining corpora do not yet exist, shifting the bottleneck from feature learning to annotation quality and dataset design.
That said, these conclusions must be regarded with appropriate caution. The Kinetics control on JIGSAWS (CCC 0.17) falls within the same range as most transfer methods (CCC 0.06 to 0.20), which further reinforces that poor JIGSAWS performance is driven by annotation-level issues rather than feature quality. However, this also means that the JIGSAWS side of the experiment cannot meaningfully contribute to validating or refuting claims about what the backbone encodes. Without reliable ground truth on at least one end of the bidirectional transfer, conclusions about the relative contributions of backbone and heads rest primarily on the LASANA results alone. While the LASANA findings are internally consistent and supported by a substantially larger and better-structured dataset, the absence of confirmatory evidence from the reverse direction limits the strength of these interpretations. A replication of this analysis on additional well-annotated datasets would be necessary to establish these conclusions more firmly.

\section{Conclusion}\label{sec:concl}
This work investigated bidirectional skill transfer between JIGSAWS and LASANA, two surgical video datasets differing in visual domain, assessment framework, and scale. The results demonstrate that cross-dataset skill transfer is feasible and effective when the target dataset is sufficiently large and well-annotated: backbones pretrained on JIGSAWS achieved CCC values of 0.77 to 0.80 on LASANA, closely matching the end-to-end baseline of 0.81. This finding supports the existence of generalizable spatiotemporal skill representations that transcend specific visual domains and assessment scales. However, transfer to JIGSAWS failed consistently across all methods, including end-to-end baselines trained directly on the dataset. This failure aligns with the statistical analysis by Hendricks et al.~\cite{Hendricks2024}, who demonstrated that GRS scores in JIGSAWS lack a consistent relationship with operator experience levels. These findings suggest that the near-decade-old gold standard warrants critical re-examination, particularly as the field moves toward more rigorous evaluation protocols that expose annotation inconsistencies previously masked by cross-validation schemes.

The control experiments using a Kinetics-pretrained backbone further revealed that surgical-domain pretraining may not be strictly necessary for skill assessment. A generic action recognition backbone achieved a CCC of 0.78 on LASANA, nearly matching all surgical-domain transfer methods. This indicates that the task-specific heads carry the majority of the skill-learning burden, while the backbone need only provide adequate spatiotemporal visual features. From a practical standpoint, this lowers the barrier to deploying skill assessment systems in new surgical domains, as costly domain-specific pretraining and end-to-end training on large surgical corpora may be replaceable by lightweight head training on well-annotated target data, reducing infrastructure requirements for clinical adoption.

Future work should validate these findings on additional well-annotated surgical skill datasets to confirm the generalizability of both the transfer approach and the head-driven skill learning observed here. Equally important is a community-level effort toward revisiting assuptions of established baselines as well as establishing larger, rigorously annotated benchmarks to support robust evaluation under held-out test set protocols, ensuring that reported advances in automated skill assessment reflect genuine predictive capability rather than artifacts of dataset structure.

\bmsubsection*{Author Contributions}
Hanna Hoffmann: Conceptualization,
Investigation, Methodology, Project Administration, Software, Validation, Formal analysis, Investigation, Writing - Original Draft, Writing - Review \& Editing, Visualization.
Felix von Bechtolsheim: Writing - Review \& Editing
Stefanie Speidel: Conceptualization, Funding acquisition, Resources, Supervision, Writing - Review \& Editing.
Rebecca Hisey: Investigation, Methodology, Software, Validation, Writing - Review \& Editing, Supervision.


\bmsubsection*{Financial Disclosure}

The authors would like to thank the Federal Ministry of Research, Technology, and Space (BMFTR) for its support as part of the research program Communication Systems “Souverän. Digital. Vernetzt.”. Joint project 6G-life, project identification number: 16KIS2413K

Funded by the German Research Foundation (DFG, Deutsche Forschungsgemeinschaft) as part of Germany's Excellence Strategy - EXC 2050/2 - Project ID 390696704 - Cluster of Excellence “Centre for Tactile Internet with Human-in-the-Loop” (CeTI) of TUD Dresden University of Technology.

\bmsubsection*{Conflicts of Interest}

No conflicts to declare.

\bibliography{wileyNJD-Chicago}





\end{document}